\documentclass{article}
\PassOptionsToPackage{final}{neurips_2026}   % camera-ready: de-anonymized, workshop footer kept

\PassOptionsToPackage{numbers, compress}{natbib}

\usepackage[dblblindworkshop]{neurips_2026}
\workshoptitle{Agentic AI Benchmarks and Applications for Enterprise Tasks}

\usepackage[utf8]{inputenc} % allow utf-8 input
\usepackage[T1]{fontenc}    % use 8-bit T1 fonts
\usepackage{hyperref}       % hyperlinks
\usepackage{url}            % simple URL typesetting
\usepackage{booktabs}       % professional-quality tables
\usepackage{amsfonts}       % blackboard math symbols
\usepackage{nicefrac}       % compact symbols for 1/2, etc.
\usepackage{microtype}      % microtypography

\usepackage{times}
\usepackage{xspace}
\usepackage{latexsym}
\usepackage{inconsolata}
\usepackage{graphicx}
\usepackage{multirow}
\usepackage{makecell}
\usepackage{colortbl}
\usepackage{xcolor}
\usepackage{amsmath,amssymb}
\usepackage{enumitem}
\usepackage{algorithm}
\usepackage{algorithmic}

\graphicspath{{./}}

\definecolor{caseRed}{RGB}{220,0,0}
\definecolor{planGreen}{RGB}{0,153,102}
\definecolor{codeTeal}{RGB}{0,153,204}
\definecolor{accentMagenta}{RGB}{204,0,153}
\definecolor{noteBlue}{RGB}{0,102,204}
\definecolor{grayText}{gray}{0.25}
\newcommand{\incode}[1]{\texttt{\textcolor{codeTeal}{#1}}}
\newcommand{\keyw}[1]{\textcolor{planGreen}{\texttt{#1}}}
\newcommand{\hilite}[1]{\textcolor{accentMagenta}{#1}}

\newcommand{\slate}{{\tt    SLATE}\xspace}

\usepackage[most]{tcolorbox}
\lstdefinelanguage{json}{
  basicstyle=\ttfamily\footnotesize,
  showstringspaces=false,
  breaklines=true,
  frame=none
}

\definecolor{promptheader}{RGB}{45,45,45}
\definecolor{promptbg}{RGB}{245,245,245}

\newtcolorbox{promptwrapper}[1]{
  enhanced,
  breakable,
  colback=gray!5,
  colframe=black!60,
  boxrule=0.6pt,
  arc=3pt,
  left=6pt,
  right=6pt,
  top=6pt,
  bottom=6pt,
  title=#1,
  coltitle=white,
  colbacktitle=black!75,
  fonttitle=\bfseries,
  attach boxed title to top left={yshift=-2mm, xshift=3mm},
  boxed title style={
    boxrule=0pt,
    arc=3pt,
    outer arc=3pt
  }
}

\newcommand{\rSmolTSd}{0.3$\pm$0.5}
\newcommand{\rSmolSAd}{52.3$\pm$0.3}
\newcommand{\rSmolTSb}{0.0$\pm$0.0}
\newcommand{\rSmolSAb}{36.6$\pm$0.2}
\newcommand{\rQwenTSd}{4.3$\pm$0.5}
\newcommand{\rQwenSAd}{72.3$\pm$0.2}
\newcommand{\rQwenTSb}{3.3$\pm$0.5}
\newcommand{\rQwenSAb}{65.7$\pm$0.6}
\newcommand{\rGemEfTSd}{0.0$\pm$0.0}
\newcommand{\rGemEfSAd}{44.1$\pm$0.5}
\newcommand{\rGemEfTSb}{1.7$\pm$0.5}
\newcommand{\rGemEfSAb}{64.4$\pm$0.3}
\newcommand{\rLlamaTSd}{1.0$\pm$0.8}
\newcommand{\rLlamaSAd}{56.1$\pm$0.4}
\newcommand{\rLlamaTSb}{0.3$\pm$0.5}
\newcommand{\rLlamaSAb}{42.0$\pm$0.6}
\newcommand{\rQwenNTSd}{\textbf{4.7$\pm$1.2}}
\newcommand{\rQwenNSAd}{\textbf{74.8$\pm$0.8}}
\newcommand{\rQwenNTSb}{2.7$\pm$0.9}
\newcommand{\rQwenNSAb}{67.7$\pm$1.3}
\newcommand{\rGemTwTSd}{1.0$\pm$0.8}
\newcommand{\rGemTwSAd}{56.6$\pm$0.9}
\newcommand{\rGemTwTSb}{\textbf{5.0$\pm$0.8}}
\newcommand{\rGemTwSAb}{\textbf{74.4$\pm$0.3}}
\newcommand{\bestTaskSuccess}{5.0\%\xspace}
\newcommand{\minStepAcc}{36.6\%\xspace}
\newcommand{\maxStepAcc}{74.8\%\xspace}
\newcommand{\maxOrderStd}{1.3\xspace}

\title{\slate: A Large-Scale Benchmark for\\ Long-Horizon API Selection in E-commerce Tool Use}

\author{
  Rongzhe Wei\thanks{Equal contribution.} \\
  Amazon \\
  \texttt{rongzhew@amazon.com}
  \And
  Ge Shi\footnotemark[1] \\
  Amazon \\
  \texttt{gegshi@amazon.com}
  \And
  Min Cheng \\
  Amazon \\
  \texttt{mmcheng@amazon.com}
  \AND
  Sarthak Ghosh \\
  Amazon \\
  \texttt{ghossart@amazon.com}
  \And
  Vaibhav Gorde \\
  Amazon \\
  \texttt{gordev@amazon.com}
  \And
  Leman Akoglu \\
  Amazon \\
  \texttt{lakoglu@amazon.com}
  \AND
  Pan Li \\
  Georgia Institute of Technology \\
  \texttt{panli@gatech.edu}
  \And
  Na Zhang \\
  Amazon \\
  \texttt{naazhang@amazon.com}
}

\begin{document}

\maketitle

\begin{abstract}
Enterprises increasingly want LLM agents to automate standard operating procedures over their software stacks---mapping each step of a procedure to the right call among thousands of internal APIs, without hand-authored step-to-API mappings. The core difficulty is \emph{API selection at scale}: choosing and composing the correct tools over a long horizon, judged only by whether the overall task succeeds. Current tool-use benchmarks do not test this regime---they expose only a handful of tools, evaluate single-step invocation, or grade multi-step behavior with subjective LLM-as-a-judge scores that penalize valid alternative trajectories. We introduce \slate (\textbf{S}ynthetic \textbf{L}arge-scale \textbf{A}PI \textbf{T}oolkit for \textbf{E}-commerce), a large-scale benchmark that tests large-toolset, long-horizon API selection in a realistic enterprise domain. \slate couples (i) queries decomposed into hierarchical, tool-grounded procedures over a $\sim$1,000-tool e-commerce library, (ii) a deterministic, context-aware simulator that returns recorded outputs for matching calls and default responses otherwise, and (iii) an end-to-end task-success metric---a task succeeds when it reaches every reference outcome---that credits diverse-but-valid trajectories while remaining fully automatic and reproducible. Experiments with six open decision models and the language models they are built on, from 3B to 12B parameters, show that even the best model completes only \bestTaskSuccess of the tasks, establishing \slate as a challenging testbed for reliable enterprise tool use. We release \slate, its simulator, and the evaluation harness.\footnote{\url{https://geshijoker.github.io/slate/}}
\end{abstract}

\section{Introduction}
\label{sec:intro}

Large language models have become capable tool-using agents that autonomously call external APIs to solve real-world tasks~\citep{yao2023react,shinn2023reflexion}, with growing potential to automate complex workflows across e-commerce~\citep{yao2022webshop,yang2024finrobot}, software development~\citep{hong2023metagpt}, and scientific discovery~\citep{boiko2023emergent}. Nowhere is this more consequential than inside enterprises. Enterprises increasingly want LLM agents to automate their \emph{standard operating procedures} (SOPs): given a procedure and a large library of internal software-service APIs---often hundreds to thousands---the agent should map each step to the right tool call and execute it, \emph{without} engineers hand-authoring a mapping from every step to a concrete API~\citep{qu2025tool,toolllm}. This demands agents that select and compose tools reliably over long horizons in large action spaces, precisely the setting of tool-augmented agents in e-commerce operations such as catalog, order, pricing, and customer-service management~\citep{yao2022webshop}. Claims of progress on such agents are only as trustworthy as the benchmarks behind them, yet existing tool-use benchmarks do not reflect this regime: small, controlled environments expose only a handful of tools~\citep{yao2024tau,shridhar2020alfworld}; large tool libraries grade single-step invocation or retrieval rather than long-horizon execution~\citep{gorilla,li2023api,tang2023toolalpaca}; and end-to-end evaluations typically fall back on subjective, non-reproducible LLM-as-a-judge scoring~\citep{toolllm,huang2024planning}. What is missing is a benchmark that tests large-toolset, long-horizon API selection in a realistic enterprise setting, where success is judged only by whether the overall task completes.

We close this gap with \slate (\textbf{S}ynthetic \textbf{L}arge-scale \textbf{A}PI \textbf{T}oolkit for \textbf{E}-commerce), a benchmark designed from the enterprise regime backward. Our contributions are:

\vspace{-1mm}
\begin{itemize}[leftmargin=1.4em,itemsep=1pt,topsep=1pt]
\item \textbf{\slate, a benchmark for large-scale API selection under end-to-end evaluation.} \slate is, to our knowledge, the first tool-use benchmark to jointly demand \emph{large tool spaces} ($\sim$1,000 tools), \emph{long-horizon} procedures, and \emph{deterministic tool outcomes}, instantiated in a realistic e-commerce domain (Fig.~\ref{fig:slate_illustration}). Its core is a \emph{deterministic, context-aware simulator} that executes every procedure without implementing thousands of live APIs, enabling an \emph{end-to-end task-success metric}: a trajectory succeeds iff it reaches every reference outcome of the task, crediting functionally equivalent solution paths while avoiding the subjectivity and false penalties of LLM-as-a-judge.
\item \textbf{A benchmarking study.} Evaluating six open decision models and the language models they are built on (3B--12B parameters) shows \slate is discriminative and that current models fall well short, providing an objective, reproducible yardstick for enterprise API selection.
\end{itemize}

\vspace{-1mm}
\section{The \slate Benchmark}
\begin{figure}[t]
  \vspace{-2mm}
  \centering
  \includegraphics[width=\columnwidth]{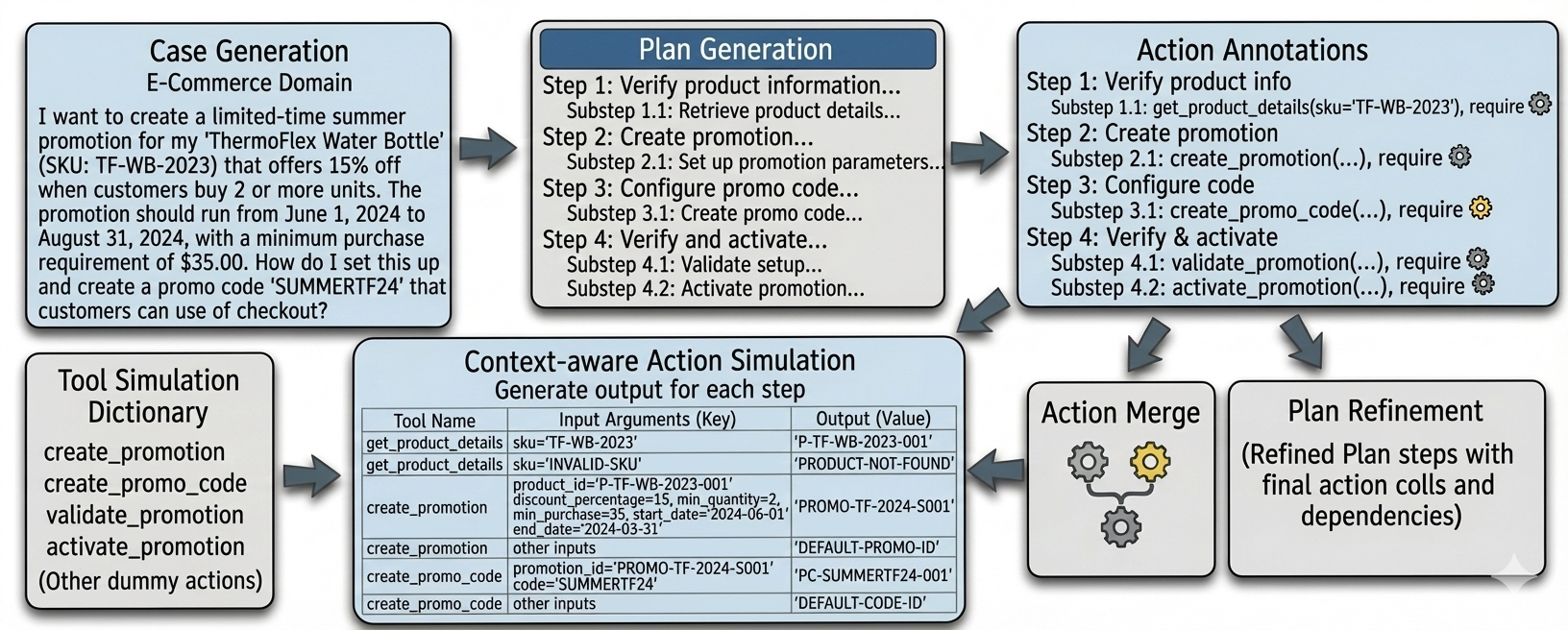}
  \caption{\textbf{Illustration of the \slate dataset structure.}}
  \label{fig:slate_illustration}
  \vspace{-6pt}
\end{figure}

\slate targets the conditions under which enterprise tool-use agents operate: selecting among a large tool library ($\sim$1,000 e-commerce APIs) to execute a long-horizon, hierarchical plan with step-wise deterministic outcomes. Each instance pairs a user query with a hierarchical solution plan whose substeps are grounded to reference tool calls, together with the simulated outputs needed to verify execution (Fig.~\ref{fig:slate_illustration}). Instances are produced by a four-stage pipeline---query generation, hierarchical plan generation, tool grounding, and context-aware simulation---followed by manual quality control; we summarize construction below and give full details, including prompt templates, in Appendix~\ref{app:slate_construction}.

Existing tool-use benchmarks typically capture only part of the challenge. Small-scale settings such as \textsc{$\tau$-Bench}~\citep{yao2024tau}, \textsc{AlfWorld}~\citep{shridhar2020alfworld}, and AppWorld~\citep{trivedi2024appworld} (457 APIs across nine apps) support controlled evaluation but do not reflect large tool spaces, while benchmarks with broader tool libraries often emphasize tool retrieval or single-step correctness rather than verifiable end-to-end execution \citep{toolllm,huang2023metatool,tang2023toolalpaca,gorilla,li2023api,huang2024planning}. Table~\ref{tab:comprehensive_comparison} organizes prior datasets along four complementary properties: \textit{Large Tool Spaces}, \textit{Long-Horizon Execution}, \textit{Deterministic Tool Outcomes}, and \textit{End-to-End Outcome Evaluation}. \slate is designed to satisfy all four simultaneously, enabling objective evaluation even when multiple execution trajectories reach the same correct final state.

\begin{table*}[ht]
\centering
\vspace{-2mm}
\caption{Comprehensive Comparison of Agent Tool-Call Datasets.}
\label{tab:comprehensive_comparison}
\resizebox{\textwidth}{!}{
\begin{tabular}{lcccc}
\toprule
\textbf{Benchmark} & \makecell{\textbf{Large Tool Spaces} \\ \textbf{($\sim$1,000 Tools)}} & \makecell{\textbf{Long-Horizon Execution} \\ \textbf{(10+ Steps)}} & \makecell{\textbf{Deterministic Tool} \\ \textbf{Outcomes}} & \makecell{\textbf{End-to-End Outcome} \\ \textbf{Evaluation}} \\
\midrule
\textbf{SLATE (Ours)} & \textbf{$\checkmark$} & \textbf{$\checkmark$} & \textbf{$\checkmark$} & \textbf{$\checkmark$} \\
TRAJECT-Bench~\citep{trajectbench} & $\checkmark$ & $\checkmark$ & $\checkmark$ & $\times$ \\
Toolathlon~\citep{toolathlon} & $\times$ & $\checkmark$ & $\checkmark$ & $\checkmark$ \\
ToolLLM~\citep{toolllm} & $\checkmark$ & $\times$ & $\checkmark$ & $\times$ \\
Gorilla~\citep{gorilla} & $\checkmark$ & $\times$ & $\checkmark$ & $\times$ \\
UltraTool~\citep{huang2024planning} & $\checkmark$ & $\times$ & $\times$ & $\times$ \\
$\tau$-bench~\citep{yao2024tau} & $\times$ & $\checkmark$ & $\times$ & $\checkmark$ \\
$\tau^2$-Bench \citep{tau2bench} & $\times$ & $\checkmark$ & $\checkmark$ & $\checkmark$ \\
AppWorld~\citep{trivedi2024appworld} & $\times$ & $\checkmark$ & $\checkmark$ & $\checkmark$ \\
ACEBench~\citep{acebench} & $\checkmark$ & $\times$ & $\checkmark$ & $\times$ \\
HammerBench~\citep{hammerbench} & $\checkmark$ & $\times$ & $\checkmark$ & $\checkmark$ \\
ToolSandbox~\citep{toolsandbox} & $\times$ & $\checkmark$ & $\times$ & $\checkmark$ \\
BFCL~\citep{bfcl} & $\times$ & $\times$ & $\checkmark$ & $\checkmark$ \\
SOP-Bench~\citep{sopbench} & $\times$ & $\checkmark$ & $\checkmark$ & $\checkmark$ \\
MetaTool~\citep{huang2023metatool} & $\times$ & $\times$ & $\times$ & $\times$ \\
API-Bank~\citep{li2023api} & $\checkmark$ & $\times$ & $\checkmark$ & $\checkmark$ \\
\bottomrule
\end{tabular}
}
\vspace{-2mm}
\end{table*}

\vspace{-1mm}
\subsection{Dataset Construction}
\vspace{-1mm}
\slate is built by a pipeline of query collection, hierarchical plan generation, tool definition and grounding, tool simulation, and manual post-processing, so that every instance is both logically sound and executable.

\begin{figure}[ht]
    \centering
    \includegraphics[width=\linewidth]{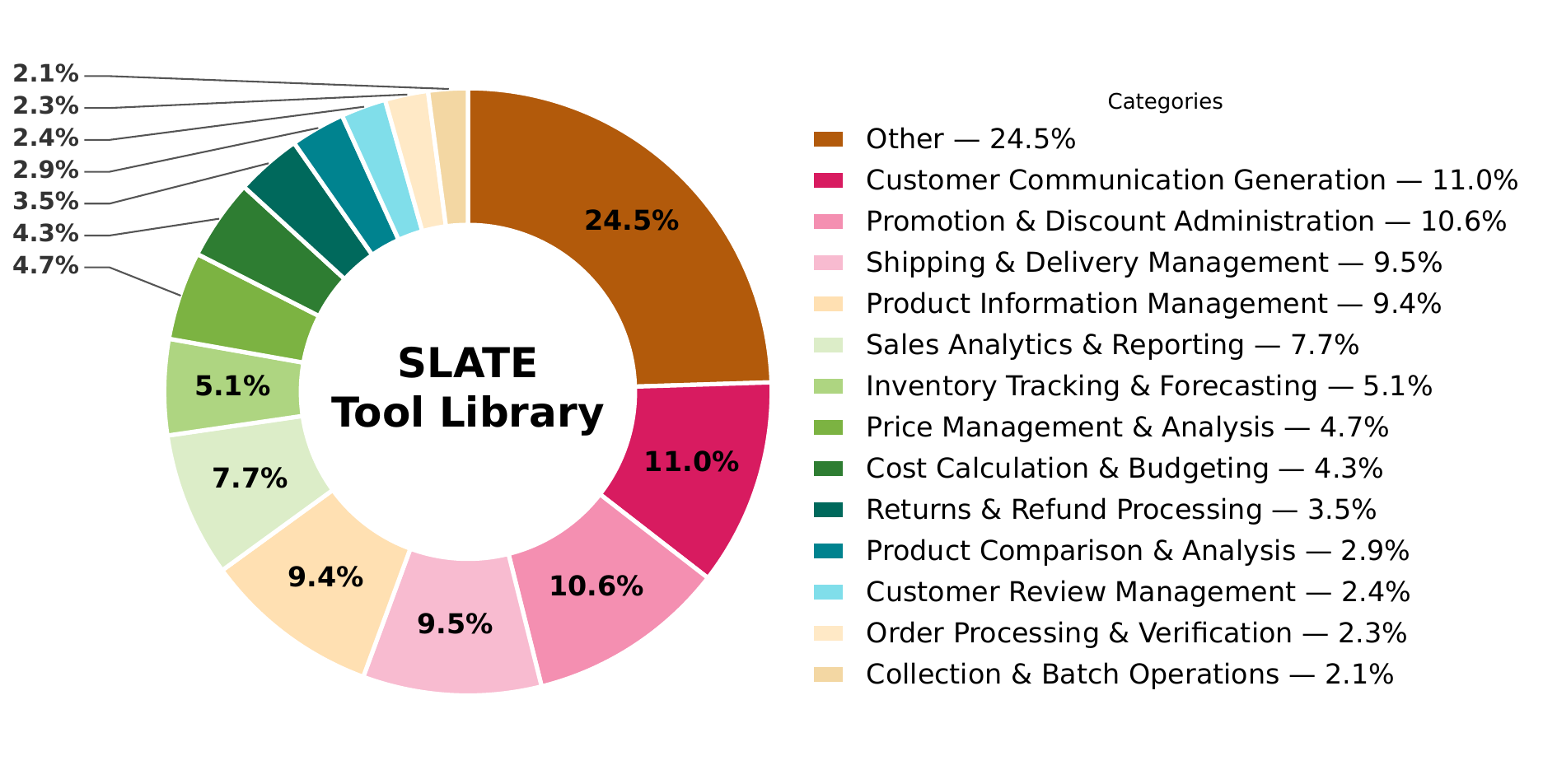}
    \vspace{-2mm}
    \caption{\textbf{SLATE tool-library category distribution.}}
    \vspace{-2mm}
    \label{fig:slate_tool_distribution}
\end{figure}

\textbf{Query collection.} We target e-commerce because its buyer--seller workflows naturally require coordinating many interdependent APIs. Working with domain experts, we define eleven representative task categories---Pre-purchase Product Questions, Product Search \& Comparison, Gifts \& Discount Codes, Order \& Delivery Changes, and Subscriptions \& Returns for buyers, and Product Information Updates, Pricing Changes, Promotions, Fulfillment \& Shipping, Inventory \& Sales Analytics, and Customer Questions \& Reviews for sellers. For each instance we sample a focus area within a category and use Claude to generate and then progressively complicate a query, yielding both linguistic variety and operational complexity. Fig.~\ref{fig:slate_tool_distribution} shows the resulting distribution of library tools across functional categories.

\textbf{Hierarchical plan generation.} Following a top-down, human-like problem-solving paradigm, each query is recursively decomposed into high-level goals, each refined into finer-grained substeps. Crucially, plan generation is performed independently of any toolset, so the decomposition reflects the natural structure of the task rather than the names of available tools, which avoids tool-specific bias in the resulting plans.

\textbf{Tool definition and grounding.} For each substep that requires a tool, we define the tool in a normalized schema specifying its name, input arguments, functionality, and output format, guided by three principles. First, tools must be specific rather than trivially mirror the substep text, so that tool selection remains non-trivial. Second, because execution plans carry strong sequential dependencies, we ensure that later tools take arguments derivable from earlier outputs or from the original query. Third, to promote reuse and avoid redundancy, we embed every tool's name and description with ModernBERT and, for each new query, retrieve the fifty most semantically related existing tools, encouraging the model to reuse them and to create new tools only when genuinely novel functionality is needed. A normalization pass then groups tools into functional categories, merges semantically equivalent ones while reconciling their input and output arguments, and updates the plans accordingly, yielding a coherent, unified plan--tool structure across the dataset.

\textbf{Tool simulator.} Because implementing a thousand live APIs is infeasible, \slate approximates tool execution with a simulator composed of two modules. A \emph{context-aware simulation database} stores ground-truth traces: during construction, each plan is annotated with its correct tool calls and argument dependencies, and Claude generates a coherent trace of simulated outputs conditioned on the query, the full plan, and prior input--output pairs, stored as a \texttt{(tool, arguments)$\rightarrow$outcome} lookup together with a default response for mismatched calls. A \emph{deterministic matcher} then maps an agent's runtime call to a stored entry beyond exact string comparison, using fixed normalization rules that absorb differences in format (e.g., date conventions) and phrasing; on a successful match it returns the recorded outcome, and otherwise the tool's default response. Every reference solution passes the simulator. This produces realistic, reproducible feedback and lets \slate measure end-to-end task success automatically. Prompt templates for every stage are provided in Appendix~\ref{app:slate_construction}.

\vspace{-1mm}
\subsection{Scale and Statistics}
\vspace{-1mm}
Table~\ref{tab:slate_stats} summarizes \slate's scale. The benchmark pairs a curated set of complex queries with a tool library of roughly one thousand APIs, of which only a small task-relevant subset is ``real'' for any given instance while the remainder act as distractors that make selection realistic. Plans are hierarchical and long-horizon, so each instance requires an agent to chain many dependent tool calls whose arguments are derived from earlier outputs. This combination -- broad tool space, deep plans, and balanced category coverage across the eleven task categories -- is what places instances in the large-toolset, long-horizon regime that motivates the benchmark.

\begin{table}[t]
\centering
\caption{Key statistics of \slate.}
\label{tab:slate_stats}
\resizebox{0.56\linewidth}{!}{
\begin{tabular}{lc}
\toprule
\textbf{Property} & \textbf{Value} \\
\midrule
\# tasks (instances)              & 100 \\
\# tools in library               & 1{,}000 \\
\# relevant tools (used in tasks) & 417 \\
\# distractor tools               & 583 \\
Avg.\ substeps per task           & 10.8 \\
Substeps per task (range)         & 7--14 \\
Avg.\ arguments per tool          & 3.07 \\
\# task categories                & 11 \\
\bottomrule
\end{tabular}
}
\vspace{-2mm}
\end{table}

\vspace{-1mm}
\section{Experiments}
\label{sec:experiments}
\vspace{-1mm}
\textbf{Setup.} The experiments test the task \slate is built for: mapping each step of a human-centric procedure to the API call that executes it. At every substep the model receives the user query, the earlier substeps with their reference tools and outputs, and the current substep, and selects one of 52 candidates, one per single-character option label (A--Z, a--z), so that the option encoding is identical across models. The candidates always include the reference tool and a ``no tool needed'' option; the other 50 are the library tools most similar to the substep under Cohere Embed v4~\citep{cohere2025embedv4}, and all 52 are listed in a fixed shuffled order. To isolate API selection, the simulator runs a correctly selected tool with its reference arguments when the steps it depends on were also selected correctly, so one wrong selection breaks every step that depends on it. We report \emph{task success}, the share of tasks that reach every \emph{reference outcome} (the recorded outputs of the final step and of every step whose output no later step uses), and \emph{step-wise accuracy}, the share of substeps with the correct selection, each averaged over three option orders.

\textbf{Models.} A \emph{decision model}, also called a System-1 model after the fast, intuitive mode of dual-process theory~\citep{kahneman2011thinking}, is a language model fine-tuned to answer multiple-choice questions directly: given a state, a question, and a list of options, it returns a calibrated probability for every option in a single forward pass instead of generating free-form text~\citep{strandsdecider2026,openjev2026,agentdecision2026,jevify2026}. We evaluate six open decision models, each paired with the backbone LLM it was trained from; the backbone LLM receives the same candidates and is read from its next-token probabilities over the option labels (Appendix~\ref{app:decision_example} shows an example). The models span five backbone families from 3B to 12B parameters; checkpoints and prompts are listed in Appendix~\ref{app:experiment_details}.

\textbf{Results.} Table~\ref{tab:model_results} shows that \slate is far from solved. Even though the reference tool is always among the candidates, step-wise accuracy ranges from \minStepAcc to \maxStepAcc, and per-step errors compound over the dependent substeps of a task. If each of a task's 10.6 tool steps were selected correctly with independent probability $p$, the task would succeed with probability $p^{10.6}$, about 4.7\% for $p = 0.75$; the best observed task success is \bestTaskSuccess. Decision models improve step-wise accuracy over their backbone LLMs in four of six pairs (by 6.6--15.7 points), whereas both Strands deciders trail their Gemma backbones (by 17.8 and 20.3 points), so decision-model training does not uniformly help on this task. Results vary little across option orders (standard deviation at most \maxOrderStd points). \slate is therefore far from saturated for small models, leaving substantial room for future work on reliable long-horizon API selection.

\begin{table}[t]
\centering
\caption{Task success and step-wise accuracy (\%) on \slate, mean $\pm$ std over three option orders. Each decision model is trained from the backbone LLM in the same row.}
\label{tab:model_results}
\resizebox{\linewidth}{!}{
\begin{tabular}{lcc|lcc}
\toprule
\textbf{Backbone LLM} & \makecell{\textbf{Task}\\\textbf{success}} & \makecell{\textbf{Step-wise}\\\textbf{accuracy}} &
\textbf{Decision model} & \makecell{\textbf{Task}\\\textbf{success}} & \makecell{\textbf{Step-wise}\\\textbf{accuracy}} \\
\midrule
SmolLM3-3B (APO) & \rSmolTSb  & \rSmolSAb  & Jevify-SmolLM3-3B   & \rSmolTSd  & \rSmolSAd \\
Qwen3-4B         & \rQwenTSb  & \rQwenSAb  & AgentDecision-4B    & \rQwenTSd  & \rQwenSAd \\
Gemma-4-E4B-it   & \rGemEfTSb & \rGemEfSAb & Strands-Decider-E4B & \rGemEfTSd & \rGemEfSAd \\
Tulu-3-8B-DPO    & \rLlamaTSb & \rLlamaSAb & Jevify-Tulu-3-8B    & \rLlamaTSd & \rLlamaSAd \\
Qwen3.5-9B       & \rQwenNTSb & \rQwenNSAb & OpenJev-Flash-9B    & \rQwenNTSd & \rQwenNSAd \\
Gemma-4-12B-it   & \rGemTwTSb & \rGemTwSAb & Strands-Decider-12B & \rGemTwTSd & \rGemTwSAd \\
\bottomrule
\end{tabular}
}
\vspace{-2mm}
\end{table}

\vspace{-1mm}
\section{Conclusion}
\label{sec:conclusion}
\vspace{-1mm}
We presented \slate, a large-scale benchmark that tests what enterprise tool-use actually demands: selecting among $\sim$1,000 APIs over a long horizon, and being graded by a deterministic, end-to-end task-success metric that accepts any valid trajectory rather than matching a single reference trace or relying on a subjective judge. Experiments with open decision models and the language models they are built on (Section~\ref{sec:experiments}) show that current models still fall well short, establishing \slate as a challenging benchmark for reliable enterprise API selection. Beyond ranking agents, \slate is intended as a testbed for improving them: a natural next step is to study different test-time-scaling strategies that spend rollout compute where it most improves end-to-end success, toward more efficient agents for enterprise API-matching problems. We release the dataset, the context-aware simulator, and the evaluation harness so the community can measure enterprise agents objectively and reproducibly, and extend the construction pipeline to new domains. By isolating API selection at scale under realistic task-level feedback, \slate provides a reproducible yardstick for enterprise tool-use agents and, we hope, a foundation for developing more reliable and compute-efficient ones. Limitations and ethical considerations are discussed in Appendix~\ref{app:limitations}.

\clearpage
\bibliographystyle{unsrtnat}
\bibliography{references}

\clearpage
\onecolumn

\begin{center}
{\Large\bfseries Appendices}
\end{center}
\vspace{0.5em}

\appendix

\section{\slate Construction}
\label{app:slate_construction}
\subsection{Prompt Templates}
\label{app:prompt_template}

\begin{figure*}[ht]
\begin{promptwrapper}{Prompt for Generating Diverse E-Commerce Cases}

Generate ONE focused e-commerce query that an \{user\_type\} might have.

User type: \{user\_type\}
Focus area: \{focus\}
Complexity level: \{complexity\}

Requirements based on COMPLEXITY LEVEL:

\{complexity\} COMPLEXITY GUIDELINES:
\{self.\_get\_complexity\_guidelines(complexity)\}

For ALL queries, ensure:
\begin{enumerate}
\item Include SPECIFIC DETAILS that can be used as direct arguments in tools:
\begin{itemize}
\item Product names and identifiers (e.g., ``Sunset Yoga Mat (SKU: YM-2023-BL)'')
\item Order numbers (e.g., ``Order \#AB-12345678'')
\item FULL DATES WITH YEARS (e.g., ``January 15, 2023'' not just ``January 15'')
\item Prices with currency (e.g., ``\$49.99'')
\item Specific quantities (e.g., ``3 units'')
\end{itemize}
\item Include MULTIPLE DATA POINTS for tools to extract as parameters
\item For COMPLEX and ADVANCED queries, include:
\begin{itemize}
\item Multiple constraints or conditions
\item Timing requirements or deadlines
\item Preferences with priorities
\item Historical context or previous actions
\item Special exceptions or unusual circumstances
\end{itemize}
\end{enumerate}

Examples of queries at different complexity levels:

SIMPLE: ``I need to track my order \#RT-78256391 for the Samsung Galaxy Buds that I ordered on May 3, 2023. When will it be delivered?''

MODERATE: ``I need to return two items from my Order \#112-9384756 placed on March 12, 2023: the Samsung Galaxy S22 with a cracked screen and the protective case. I want to keep the screen protector and charging cable. Can I get a return label for just those two items?''

COMPLEX: ``I need to modify my bulk order \#BLK-2023-4872 placed on February 28, 2023 for my company. We ordered 50 Lenovo ThinkPad T14 laptops with i7 processors at \$1,299 each, but now I need to change 15 of them to the i9 model which costs \$1,599 each. We've already paid the deposit of \$25,000 and delivery is scheduled for June 15, 2023. I need to know if this change will affect our delivery date and what additional payment is required.''

\end{promptwrapper}
\caption{Prompt to generate diverse e-commerce cases from different user perspectives. \textbf{Example Variable Values:} \texttt{user\_type} alternates between ``buyer'' and ``seller'' based on query ID; \texttt{focus} is randomly selected from buyer focus areas (``product search with multiple conflicting requirements'', ``order tracking for multiple items with shipping complications'', ``complex product return with partial refund request'', etc.) or seller focus areas (``updating complex product variations and attributes'', ``inventory management across multiple warehouses'', ``implementing tiered pricing strategy with conditions'', etc.); \texttt{complexity} is one of ``SIMPLE'', ``MODERATE'', ``COMPLEX'', or ``ADVANCED'' with weighted distribution of 0\%, 10\%, 80\%, 10\% respectively.}
\end{figure*}
\clearpage
\begin{figure*}[ht]
\begin{promptwrapper}{Prompt for Creating Specialized Tool Plans to Resolve E-Commerce Queries (Part 1)}

ADVANCED: ``I'm managing our company's quarterly office supply order (PO \#BZ-45721) placed on April 2, 2023 with scheduled delivery on April 20, 2023 across three locations. For the Chicago office (Location ID: CHI-005), we need to cancel the 12 ergonomic chairs (\$259 each) due to their recent merger, but expedite the 15 monitor stands (\$89 each) to arrive by April 15. For the Boston office (Location ID: BOS-002), we need to add 8 wireless keyboards (\$65 each) and 8 wireless mice (\$45 each) for new hires. The New York office (Location ID: NYC-001) shipment is fine as is, but we need to change the delivery window to after 2:00 PM. We're eligible for the 12\% corporate discount and already applied the SPRING2023 promo code for 5\% off. How will these changes affect our total, and can all these modifications be accommodated before processing begins on April 10?''

Return only the query text with no additional explanation.

Create a plan with specialized tools to resolve this e-commerce query:

User Query (\{user\_type\}, \{complexity\} complexity): ``\{query\}''

\{existing\_tools\_text\}

Please provide:

1. A HIERARCHICAL PLAN with:
\{plan\_requirements\}

2. For each SUBSTEP, indicate:
\begin{itemize}
\item If it's a high-level step: ``tool'': ``null'' - NO EXCEPTIONS!
\item If no tool is needed: ``tool'': ``No tool required''
\item If a tool is needed: ``tool'': ``tool\_name(param1='value1', param2='value2')''
\end{itemize}

3. CRITICAL: FIRST TOOL CALL arguments must come DIRECTLY from the query
\begin{itemize}
\item Example: If query mentions ``Order \#AB-12345'', first tool should use order\_id='AB-12345'
\item Extract actual values from the query text, don't invent new values
\end{itemize}

4. CRITICAL: SUBSTEP DESCRIPTIONS MUST BE DISTINCT FROM TOOL NAMES
\begin{itemize}
\item Make each substep description MEANINGFUL and CONTEXT-RICH
\item Do not create tool names that are Identical or Very Similar to substep descriptions.
\item BAD: ``1.1 Get order details'' when using ``get\_order\_details'' tool
\item GOOD: ``1.1 Retrieve customer's purchase history for Order \#AB-123'' when using ``get\_order\_details'' tool
\item Describe the PURPOSE and CONTEXT of the step, not just the action
\item Include relevant business context and specific goals for each step
\end{itemize}

\end{promptwrapper}
\caption{Prompt for creating specialized tool plans (Part 1). \textbf{Example Variable Values:} \texttt{user\_type} is either ``buyer'' or ``seller''; \texttt{complexity} is one of ``SIMPLE'', ``MODERATE'', ``COMPLEX'', or ``ADVANCED''; \texttt{query} is the actual query text generated in stage 1. \textbf{existing\_tools\_text:} Dynamically generated list of top 30 most relevant tools from global tool library. \textbf{plan\_requirements:} SIMPLE (2-3 high-level steps, 1-2 substeps each, 3-5 total); MODERATE (3-4 high-level steps, 1-3 substeps each, 6-8 total); COMPLEX (4-5 high-level steps, 2-3 substeps each, 8-12 total); ADVANCED (5-7 high-level steps, 2-4 substeps each, 12-16 total).}
\end{figure*}

\begin{figure*}[ht]
\begin{promptwrapper}{Prompt for Creating Specialized Tool Plans to Resolve E-Commerce Queries (Part 2)}

5. TOOL SELECTION AND DESIGN:
\begin{itemize}
\item TRY REUSE EXISTING TOOLS whenever appropriate (from the list above)
\item If none of the existing tools are suitable, design NEW SPECIALIZED TOOLS that:
\begin{itemize}
\item Have SPECIFIC, FOCUSED functionality (not general-purpose)
\item Use SIMPLE arguments (2-3 parameters maximum)
\item Return SIMPLE, FOCUSED results (1-3 fields maximum)
\item Follow snake\_case naming convention
\item Avoid creating monolithic ``do everything'' tools
\end{itemize}
\end{itemize}

6. CRITICAL - AVOID TOOL REPETITION:
\begin{itemize}
\item DO NOT use the same tool in consecutive substeps (e.g., avoid ``2.1 xxx tool\_A; 2.2 xxx tool\_A'')
\item If you need to call the same tool multiple times, space them out with other operations
\item Each substep should ideally use a DIFFERENT tool to create diversity
\item Create specialized tools for different aspects rather than reusing generic ones
\end{itemize}

7. IMPORTANT - SEQUENTIAL DEPENDENCIES:
\begin{itemize}
\item Later steps should use results from previous steps for sequential dependencies
\item Use the format OUTPUT\_FROM\_STEP\_X.Y.field to reference previous outputs
\item Example: product\_id=OUTPUT\_FROM\_STEP\_1.2.product\_id
\item Create a CHAIN of dependencies where each step builds on previous results
\item Ensure that tool outputs from early steps provide necessary inputs for later steps
\end{itemize}

8. FINAL STEP should produce a DIRECT RESULT that resolves the query
\begin{itemize}
\item The last tool should return a clear outcome or answer
\item For example: confirmation message, success status, or direct result
\item FINAL STEPS should utilize outputs from earlier steps
\end{itemize}

9. FOR \{complexity\} COMPLEXITY:
\begin{itemize}
\item \{self.\_get\_plan\_complexity\_guidelines(complexity)\}
\end{itemize}

\end{promptwrapper}
\caption{Prompt for creating specialized tool plans (Part 2). Focus on tool selection, reuse strategy, avoiding repetition, and establishing sequential dependencies between steps.}
\end{figure*}

\begin{figure*}[ht]
\begin{promptwrapper}{Prompt for Creating Specialized Tool Plans to Resolve E-Commerce Queries (Part 3)}

10. TOOL DIVERSITY REQUIREMENTS:
\begin{itemize}
\item Create tools that span different functional domains (retrieval, validation, processing, notification, etc.)
\item Avoid generic tools like ``process\_request'' or ``handle\_query''
\item Instead create specific tools like ``validate\_return\_window'', ``calculate\_refund\_amount'', ``send\_confirmation\_email''
\item Each tool should have a clear, single responsibility
\end{itemize}

\medskip
\textbf{TOOL DESIGN PRINCIPLES:}
\begin{enumerate}
\item FOCUSED: Each tool should do ONE thing well
\item COMPOSABLE: Tools should work together through their inputs/outputs
\item REUSABLE: Create tools that could be useful in other scenarios
\item SIMPLE: Prefer multiple simple tools over one complex tool
\item DIVERSE: Create tools spanning different functional domains
\item SEQUENTIAL: Design tools to create natural dependencies and data flow
\end{enumerate}

\medskip
\textbf{Remember:}
\begin{itemize}
\item REUSE existing tools whenever appropriate
\item FIRST TOOL must use arguments DIRECTLY from the query
\item Keep tools SPECIALIZED with SIMPLE inputs and outputs
\item Ensure steps are SEQUENTIAL with clear dependencies
\item AVOID repeating the same tool in consecutive steps
\item FINAL STEP should provide a DIRECT RESOLUTION to the query
\item Return valid JSON only
\end{itemize}

\end{promptwrapper}
\caption{Prompt for creating specialized tool plans (Part 3). Format response as JSON object with ``plan'' array (high-level steps with ``tool'': ``null'', and substeps with tool calls or ``No tool required'') and ``tools'' array (defining all tools with ``name'', ``description'', ``arguments'', and ``results'' fields). \textbf{Complexity Guidelines:} SIMPLE (1-2 tool calls, linear, minimal dependencies); MODERATE (2-4 tool calls, at least one dependency and conditional step); COMPLEX (4-7 tool calls, multiple dependencies, validation, branching paths); ADVANCED (7+ tool calls, multi-stage workflow, complex dependencies, edge case handling, maximum tool diversity).}
\end{figure*}
\clearpage
\begin{figure*}[ht]
\begin{promptwrapper}{Prompt for Generating Diverse E-Commerce Tools}

Generate \{batch\_size\} DIVERSE e-commerce tools with COMPLETELY UNIQUE NAMES.

\medskip
\textbf{TOOL NAMING REQUIREMENTS:}
\begin{enumerate}
\item Each tool name MUST be UNIQUE and use snake\_case format
\item NEVER use any of these existing names: \{names\_str\}
\item Create DISTINCTIVE names that avoid generic patterns
\item Consider these name patterns for inspiration: \{name\_pattern\_text\}
\item Each tool name should reflect its SPECIFIC FUNCTION and DOMAIN
\item VERIFY that each name is different from all others in your response
\end{enumerate}

\medskip
\textbf{TOOL ASSIGNMENTS} - Create exactly one tool for EACH of these specific domains:
\{domains\_text\}

\medskip
\textbf{SAMPLE TOOL STRUCTURE:}
\{json.dumps(sample\_tool, indent=2)\}

\medskip
\textbf{TOOL REQUIREMENTS:}
\begin{enumerate}
\item Match each tool precisely to its assigned domain above
\item Include 1-3 SIMPLE arguments with clear purposes
\item Return 1-3 FOCUSED result fields
\item Include ``query\_id'': ``dummy'' in each tool
\item Follow the exact JSON structure of the sample
\end{enumerate}

\medskip
Return a JSON array containing \{batch\_size\} tools with UNIQUE names:
\begin{verbatim}
[
  {tool1},
  {tool2},
  ...
]
\end{verbatim}

ONLY return the JSON array with no additional text.

\end{promptwrapper}
\caption{Prompt for generating diverse e-commerce tools with unique names. \textbf{Example Variable Values:} \texttt{batch\_size} is configurable (typically 10 or 20, default 10). \texttt{names\_str} is a comma-separated string of 20 random existing tool names to avoid duplication (e.g., ``get\_order\_details, validate\_return\_window, calculate\_refund, search\_products, update\_inventory, generate\_shipping\_label, process\_payment, send\_notification, track\_shipment, analyze\_sales, manage\_promotions, verify\_address, calculate\_tax, check\_stock, create\_invoice, update\_customer, generate\_report, schedule\_delivery, process\_return, validate\_coupon''). \texttt{name\_pattern\_text} consists of 4 randomly selected naming patterns from: ``feature\_specific\_domain, domain\_specific\_action, specialized\_task\_handler, domain\_analyzer, action\_target\_generator, domain\_insight\_provider, specialized\_workflow\_automation, target\_specific\_optimizer''. \texttt{domains\_text} is a numbered list of specific domains (one per tool) from 100 predefined diverse domains covering: Product management (catalog, variants, bundling, photography assessment, competitive analysis, recommendations, sourcing, limited editions, warranties, etc.), Shipping \& logistics (delivery, warehousing, returns, tracking, etc.), etc. \texttt{sample\_tool} is a randomly selected existing tool or default structure containing: \texttt{name}, \texttt{description}, \texttt{query\_id}, \texttt{arguments} (with type, properties, and argument details), and \texttt{results} (with type, properties, and result field details).}
\end{figure*}
\clearpage
\begin{figure*}[ht]
\begin{promptwrapper}{Prompt for Categorizing E-Commerce Tools into Functional Domains}

Analyze these e-commerce tools and categorize them into functional domains.

\medskip
\textbf{TOOLS TO CATEGORIZE:}
\{json.dumps(batch\_tools, indent=2)\}

\medskip
\textbf{PREDEFINED CATEGORIES:}
\{json.dumps(self.categories, indent=2)\}

\medskip
\textbf{INSTRUCTIONS:}
\begin{enumerate}
\item For each tool, assign it to the MOST APPROPRIATE category from the predefined list
\item If a tool could fit multiple categories, choose the PRIMARY function
\item Consider the tool's main purpose and typical use case
\item Return a JSON object mapping tool names to their categories
\end{enumerate}

\medskip
\textbf{Return format:}
\begin{verbatim}
{
  "tool_name_1": "Category Name",
  "tool_name_2": "Category Name",
  ...
}
\end{verbatim}

ONLY return the JSON object with no additional text.

\end{promptwrapper}
\caption{Prompt for categorizing e-commerce tools into functional domains. \textbf{Example Variable Values:} \texttt{batch\_tools} is a batch of 10-20 tools to categorize, each containing \texttt{name}, \texttt{description}, \texttt{arguments}, and \texttt{results} fields (e.g., ``get\_order\_details'', ``calculate\_shipping\_cost'', etc.). \texttt{self.categories} is a predefined list of 14 functional categories: ``Product Information Management'', ``Shipping \& Delivery Management'', ``Promotion \& Discount Administration'', ``Sales Analytics \& Reporting'', ``Price Management \& Analysis'', ``Inventory Tracking \& Forecasting'', ``Customer Communication Generation'', ``Order Processing \& Verification'', ``Product Comparison \& Analysis'', ``Collection \& Batch Operations'', ``Returns \& Refund Processing'', ``Cost Calculation \& Budgeting'', ``Customer Review Management'', and ``Other''.}
\end{figure*}
\clearpage
\begin{figure*}[ht]
\begin{promptwrapper}{Prompt for Generating Tool Simulation Outputs}

Generate a realistic JSON output for this e-commerce tool call.

QUERY:
``\{context['query']\}''

CURRENT STEP:
``\{context['step']\}''

TOOL:
\begin{itemize}
\item Name: \{tool\_name\}
\item Description: \{tool\_def.get('description', 'No description available')\}
\end{itemize}

ARGUMENTS USED:
\{json.dumps(args, indent=2)\}

PREVIOUS TOOL OUTPUTS:
\{json.dumps(context['previous\_outputs'], indent=2) if context['previous\_outputs'] else ``No previous outputs''\}

REQUIRED OUTPUT PROPERTIES:
\{json.dumps(result\_props, indent=2)\}

EXAMPLE FORMAT:
\{json.dumps(example\_output, indent=2)\}

COMPLEXITY LEVEL: \{complexity\}
\{complexity\_guidance\}

INSTRUCTIONS:
\begin{enumerate}
\item Generate SPECIFIC, REALISTIC values consistent with the original query
\item Maintain CONSISTENCY with previous tool outputs
\item Include ALL required properties in the tool's result schema
\item Values should match their expected data types
\item \{``This is the FINAL TOOL in the plan. Make sure the output provides a DIRECT RESULT that clearly resolves the user's request (e.g., confirmation, status, or answer)'' if is\_final\_step else ``Keep output focused and relevant to the step''\}
\end{enumerate}

RETURN ONLY THE JSON OUTPUT OBJECT.

\end{promptwrapper}
\caption{Prompt for generating tool simulation outputs. \textbf{Example Variable Values:} \texttt{context['query']} is the original e-commerce query text; \texttt{context['step']} is the current step description from the plan; \texttt{tool\_name} is the name of the tool being called; \texttt{tool\_def} is the tool definition object containing description and other metadata; \texttt{args} is a JSON object containing the arguments passed to the tool; \texttt{context['previous\_outputs']} is a JSON object or array containing outputs from previously executed tools (null or empty if no previous outputs); \texttt{result\_props} is a JSON schema object defining required output properties with types and descriptions; \texttt{example\_output} is a sample JSON object showing the expected output format; \texttt{complexity} is one of ``SIMPLE'', ``MODERATE'', ``COMPLEX'', or ``ADVANCED''; \texttt{complexity\_guidance} is text providing specific guidance for generating outputs at the given complexity level; \texttt{is\_final\_step} is a boolean indicating whether this tool is the final step in the plan (affects instruction 5).}
\end{figure*}
\clearpage

\subsection{An Example of \slate Data}
\label{sec:app_examples}
This section presents an illustrative example in the \slate format, illustrating the transformation from a user query to a hierarchical execution plan, associated tool calls, and deterministic simulation outputs. The example demonstrates the procedure structure and execution semantics used for step-wise and end-to-end evaluation.

\begin{figure*}[ht]
\begin{promptwrapper}{Case}
\noindent
I want to create a limited-time summer promotion for my
\hilite{\textbf{``ThermoFlex Water Bottle''}} \;
(\textbf{SKU:} \incode{TF-WB-2023})
that offers \textbf{15\% off} when customers \textbf{buy 2 or more units}.
The promotion should run from \textbf{June 1, 2024} to \textbf{August 31, 2024},
with a minimum purchase requirement of \textbf{\$35.00}.
How do I set this up and create a promo code \hilite{\textbf{``SUMMERTF24''}}
that customers can use at checkout?
\end{promptwrapper}

\begin{promptwrapper}{Plan}
\begin{itemize}[leftmargin=1.2em,itemsep=2pt,topsep=2pt]
  \item \textcolor{grayText}{\textbf{step}}: \textcolor{grayText}{1. Verify product information}
  \begin{itemize}[leftmargin=1.4em,itemsep=1pt,topsep=1pt]
    \item \textcolor{grayText}{\textbf{step}}: \textcolor{grayText}{1.1 Retrieve product details}
    \item \textcolor{grayText}{\textbf{action}}: \keyw{get\_product\_details}\,(\textbf{sku}=\incode{``TF-WB-2023''})
  \end{itemize}

  \item \textcolor{grayText}{\textbf{step}}: \textcolor{grayText}{2. Create the promotion}
  \begin{itemize}[leftmargin=1.4em,itemsep=1pt,topsep=1pt]
    \item \textcolor{grayText}{\textbf{step}}: \textcolor{grayText}{2.1 Set up promotion parameters}
    \item \textcolor{grayText}{\textbf{action}}:
    \keyw{create\_promotion}\,(
      \textbf{product\_id}=\incode{OUTPUT\_FROM\_STEP\_1.1.product\_id},
      \textbf{discount\_percentage}=15,
      \textbf{min\_quantity}=2,
      \textbf{min\_purchase}=35.00,
      \textbf{start\_date}=\incode{``2024-06-01''},
      \textbf{end\_date}=\incode{``2024-08-31''}
    )
  \end{itemize}

  \item \textcolor{grayText}{\textbf{step}}: \textcolor{grayText}{3. Configure promo code}
  \begin{itemize}[leftmargin=1.4em,itemsep=1pt,topsep=1pt]
    \item \textcolor{grayText}{\textbf{step}}: \textcolor{grayText}{3.1 Create promo code}
    \item \textcolor{grayText}{\textbf{action}}:
    \keyw{create\_promo\_code}\,(
      \textbf{promotion\_id}=\incode{OUTPUT\_FROM\_STEP\_2.1.promotion\_id},
      \textbf{code}=\hilite{\texttt{``SUMMERTF24''}}
    )
  \end{itemize}

  \item \textcolor{grayText}{\textbf{step}}: \textcolor{grayText}{``4. Verify and activate the promotion''}
  \begin{itemize}[leftmargin=1.4em,itemsep=1pt,topsep=1pt]
    \item \textcolor{grayText}{\textbf{step}}: \textcolor{grayText}{``4.1 Validate promotion setup''}
    \item \textcolor{grayText}{\textbf{action}}:
    \keyw{validate\_promotion}\,(
      \textbf{promotion\_id}=\incode{OUTPUT\_FROM\_STEP\_2.1.promotion\_id}
    )
    \item \textcolor{grayText}{\textbf{step}}: \textcolor{grayText}{``4.2 Activate the promotion''}
    \item \textcolor{grayText}{\textbf{action}}:
    \keyw{activate\_promotion}\,(
      \textbf{promotion\_id}=\incode{OUTPUT\_FROM\_STEP\_2.1.promotion\_id},
      \textbf{promo\_code\_id}=\incode{OUTPUT\_FROM\_STEP\_3.1.promo\_code\_id}
    )
  \end{itemize}
\end{itemize}
\end{promptwrapper}
\caption{Example of plan generation and tool execution.}
\end{figure*}

\begin{figure*}[ht]

\begin{promptwrapper}{Simulate Outputs with context}
\begin{itemize}[leftmargin=1.2em,itemsep=3pt,topsep=2pt]
  \item \keyw{get\_product\_details}\,(\textbf{sku}=\incode{``TF-WB-2023''}) \\
  \textcolor{noteBlue}{$\Rightarrow$ Output:} \textbf{product\_id}=\incode{``P-TF-WB-2023-001''}

  \item \keyw{create\_promotion}\,(
    \textbf{product\_id}=\incode{``P-TF-WB-2023-001''},
    \textbf{discount\_percentage}=15,
    \textbf{min\_quantity}=2,
    \textbf{min\_purchase}=35.00,
    \textbf{start\_date}=\incode{``2024-06-01''},
    \textbf{end\_date}=\incode{``2024-08-31''}
  ) \\
  \textcolor{noteBlue}{$\Rightarrow$ Output:} \textbf{promotion\_id}=\incode{``PROMO-TF-2024-S001''}

  \item \keyw{create\_promo\_code}\,(
    \textbf{promotion\_id}=\incode{``PROMO-TF-2024-S001''},
    \textbf{code}=\hilite{\texttt{``SUMMERTF24''}}
  ) \\
  \textcolor{noteBlue}{$\Rightarrow$ Output:} \textbf{promo\_code\_id}=\incode{``PC-SUMMERTF24-001''}

  \item \keyw{validate\_promotion}\,(\textbf{promotion\_id}=\incode{``PROMO-TF-2024-S001''})

  \item \keyw{activate\_promotion}\,(
    \textbf{promotion\_id}=\incode{``PROMO-TF-2024-S001''},
    \textbf{promo\_code\_id}=\incode{``PC-SUMMERTF24-001''}
  ) \\
  \textcolor{noteBlue}{$\Rightarrow$ Output:} \textbf{success}=\incode{``true''}
\end{itemize}
\end{promptwrapper}

\begin{promptwrapper}{Deterministic Simulation Dictionary}
\noindent \textcolor{noteBlue}{\textbf{1.}} \keyw{get\_product\_details} \; \textcolor{grayText}{(Args: sku)} \\
\hspace*{1.2em}\incode{TF-WB-2023} \;\; \textcolor{noteBlue}{$\Rightarrow$}\;\; \incode{P-TF-WB-2023-001} \\
\hspace*{1.2em}\textcolor{grayText}{Other value} \;\; \textcolor{noteBlue}{$\Rightarrow$}\;\; \textcolor{grayText}{different default value}

\vspace{4pt}
\noindent \textcolor{noteBlue}{\textbf{2.}} \keyw{create\_promotion} \; \textcolor{grayText}{(Args: product\_id, discount\_percentage, min\_quantity, min\_purchase, start\_date, end\_date)} \\
\hspace*{1.2em}\incode{P-TF-WB-2023-001, 15, 2, 35, 2024-06-01, 2024-08-31}
\;\; \textcolor{noteBlue}{$\Rightarrow$}\;\; \incode{PROMO-TF-2024-S001} \\
\hspace*{1.2em}\textcolor{grayText}{Other value} \;\; \textcolor{noteBlue}{$\Rightarrow$}\;\; \textcolor{grayText}{different default value}

\vspace{4pt}
\noindent \textcolor{noteBlue}{\textbf{3.}} \keyw{create\_promo\_code} \; \textcolor{grayText}{(Args: promotion\_id, code)} \\
\hspace*{1.2em}\incode{PROMO-TF-2024-S001, SUMMERTF24}
\;\; \textcolor{noteBlue}{$\Rightarrow$}\;\; \incode{PC-SUMMERTF24-001} \\
\hspace*{1.2em}\textcolor{grayText}{Other value} \;\; \textcolor{noteBlue}{$\Rightarrow$}\;\; \textcolor{grayText}{different default value}
\end{promptwrapper}
\caption{Deterministic simulation dictionary for tool calls.}
\end{figure*}

\clearpage

\section{Experimental Details}
\label{app:experiment_details}

\textbf{Candidates.} Every candidate list contains 52 options, one per single-character label (A--Z, then a--z), which keeps the option encoding identical across models. Two options are always present: the reference tool and a ``\texttt{no\_tool\_needed}'' option. The other 50 are retrieved with Cohere Embed v4 (\texttt{cohere.embed-v4}): library tools are ranked by the cosine similarity between the embedding of ``\texttt{name: description}'' and the embedding of the user query concatenated with the substep text, and the 50 highest-ranked tools other than the reference tool are kept. The list is shuffled with a seed derived from the task and substep identifiers, and the same order is shown to a decision model and to its backbone LLM; two further orders are drawn with different seeds. Because the reference tool is always included, retrieval does not cap accuracy in these experiments. Without this guarantee, the 51 top-ranked tools would contain the reference tool for 91.1\% of tool substeps, and for every tool substep of a task in only 48\% of tasks.

\textbf{Context.} The state lists the user query and every earlier substep with its reference tool and simulated output. Each earlier output is truncated to 400 characters, so that every prompt fits in a 4,096-token window.

\textbf{Scoring.} The prerequisites of a substep are the earlier tool steps whose outputs its reference arguments use. A correctly selected tool counts as executed only when all of its prerequisites were executed. A task succeeds when every reference outcome is reached, i.e., the recorded outputs of the final tool step and of every tool step whose output no later step uses.

\textbf{Readouts.} Each decision model uses its own published inference recipe (prompt format and readout) on the shared state, step, and candidates. Each backbone LLM receives the prompt below through its chat template with thinking disabled, and its prediction is the label (A--Z, a--z) with the highest next-token probability. All checkpoints are pinned to the revisions in Table~\ref{tab:checkpoints} and run in bf16 with Hugging Face Transformers.

\begin{table}[ht]
\centering
\caption{Checkpoints (Hugging Face repositories and revisions). Each decision model is trained from the backbone LLM in its row.}
\label{tab:checkpoints}
\resizebox{\linewidth}{!}{
\begin{tabular}{lll}
\toprule
\textbf{Decision model} & \textbf{Backbone LLM} & \textbf{Readout} \\
\midrule
\texttt{Praveenrajus/jevify-smollm3-3b-apo-readout} @\texttt{553e4965} & \texttt{HuggingFaceTB/SmolLM3-3B-checkpoints} @\texttt{cfb32d50} & label probabilities (LoRA)~\citep{jevify2026} \\
\texttt{JevAgentHarness/AgentDecision-4B} @\texttt{f0e13bab} & \texttt{Qwen/Qwen3-4B} @\texttt{1cfa9a72} & label probabilities (LoRA)~\citep{agentdecision2026} \\
\texttt{StrandsAgents/strands-decider-E4B-gemma4-v1-2610} @\texttt{1344d443} & \texttt{google/gemma-4-E4B-it} @\texttt{ee0ef602} & pointer head (LoRA)~\citep{strandsdecider2026} \\
\texttt{Praveenrajus/Llama-3.1-Tulu-3-8B-DPO-jevify-readout} @\texttt{3918311a} & \texttt{allenai/Llama-3.1-Tulu-3-8B-DPO} @\texttt{a7beb67e} & label probabilities (LoRA)~\citep{jevify2026} \\
\texttt{openjev/OpenJev-Flash-9B} @\texttt{b77b0fff} & \texttt{Qwen/Qwen3.5-9B} @\texttt{c2022362} & label probabilities (merged)~\citep{openjev2026} \\
\texttt{StrandsAgents/strands-decider-12B-gemma4-v1-2610} @\texttt{1202634a} & \texttt{google/gemma-4-12B-it} @\texttt{43c9d014} & pointer head (LoRA)~\citep{strandsdecider2026} \\
\bottomrule
\end{tabular}
}
\end{table}

\textbf{Decision models versus LLMs: an example.}
\label{app:decision_example}
A regular LLM answers by generating text, which must then be parsed into a choice. A decision model instead scores the given options and returns a probability distribution over them in one forward pass. To compare the two on equal terms, we read the backbone LLM's next-token probabilities over the option labels. Figure~\ref{fig:decision_example} shows one logged decision from our runs.

\begin{figure}[ht]
\begin{promptwrapper}{{One decision: OpenJev-Flash-9B and its backbone LLM (Qwen3.5-9B)}}
\small
\textbf{State.} TASK: I need to add gift wrapping to my order \#RT-78296431 \ldots{} for the Bose QuietComfort Ultra Headphones (SKU: BQ-2023-BLK) \ldots{} Can I add a personalized message and choose a holiday-themed wrapping paper before it ships tomorrow? \ldots{} \quad HISTORY: (none)

\textbf{Current step.} 1.1 Check if gift wrapping is available for the headphones

\textbf{Options} (4 of 52 shown).\\
B. check\_gift\_wrapping\_availability -- Check if gift wrapping is available for a specific product\\
S. validate\_note\_character\_limit -- Validate if requested note length is within allowed limits\\
p. add\_gift\_wrapping\_to\_order -- Add gift wrapping and personalization options to an existing order\\
a. no\_tool\_needed -- No tool call is needed for this step.

\medskip
\textbf{Decision model} (OpenJev-Flash-9B), probabilities over the 52 options: B 0.873, Z 0.030, S 0.023, \ldots{} $\Rightarrow$ B (correct).\\
\textbf{Backbone LLM} (Qwen3.5-9B), next-token probabilities over the labels: p 0.444, B 0.346, b 0.127, \ldots{} $\Rightarrow$ p (incorrect: it adds the wrapping instead of first checking availability).
\end{promptwrapper}
\caption{A decision from the experiments (task 177, step 1.1, first option order). Z and b are \texttt{generate\_gift\_wrapping\_confirmation} and \texttt{append\_express\_upgrade\_followups}.}
\label{fig:decision_example}
\end{figure}

\textbf{Error analysis.} Table~\ref{tab:errors} breaks errors down by type. Nearly all errors select a wrong tool; skipping a needed tool or calling a tool for a no-tool substep is rare. Accuracy on substeps whose description names the reference tool is similar to or lower than on the other substeps. Table~\ref{tab:categories} reports results by task category.

\begin{table}[ht]
\centering
\caption{Error types and tool-step accuracy (\%), pooled over three option orders. Error columns are shares of all substeps:
a wrong tool selected, no tool selected for a tool step, or a tool selected for a no-tool step. The last two columns split tool-step
accuracy by whether the substep text names its reference tool (16.1\% of tool steps do).}
\label{tab:errors}
\resizebox{\linewidth}{!}{
\begin{tabular}{llccccc}
\toprule
& & \multicolumn{3}{c}{\textbf{Errors (\% of substeps)}} & \multicolumn{2}{c}{\textbf{Tool-step accuracy}} \\
\cmidrule(lr){3-5}\cmidrule(lr){6-7}
\textbf{Model} & \textbf{Type} & \makecell{\textbf{Wrong}\\\textbf{tool}} & \makecell{\textbf{Missed}\\\textbf{tool}} & \makecell{\textbf{Spurious}\\\textbf{tool}} & \makecell{\textbf{Step names}\\\textbf{its tool}} & \makecell{\textbf{Other}\\\textbf{steps}} \\
\midrule
SmolLM3-3B (APO) & Backbone LLM & 58.1 & 3.5 & 1.8 & 36.3 & 37.4 \\
Jevify-SmolLM3-3B & Decision model & 44.5 & 1.3 & 2.0 & 53.0 & 53.4 \\
\midrule
Qwen3-4B & Backbone LLM & 31.3 & 1.1 & 1.8 & 66.7 & 66.9 \\
AgentDecision-4B & Decision model & 26.0 & 0.0 & 1.7 & 73.9 & 73.3 \\
\midrule
Gemma-4-E4B-it & Backbone LLM & 33.5 & 0.2 & 1.9 & 53.2 & 67.9 \\
Strands-Decider-E4B & Decision model & 53.9 & 0.0 & 2.0 & 26.3 & 48.6 \\
\midrule
Tulu-3-8B-DPO & Backbone LLM & 55.7 & 0.3 & 2.0 & 42.1 & 42.9 \\
Jevify-Tulu-3-8B & Decision model & 37.4 & 4.5 & 2.0 & 54.2 & 57.8 \\
\midrule
Qwen3.5-9B & Backbone LLM & 28.2 & 2.5 & 1.7 & 65.1 & 69.4 \\
OpenJev-Flash-9B & Decision model & 22.7 & 0.7 & 1.8 & 73.9 & 76.5 \\
\midrule
Gemma-4-12B-it & Backbone LLM & 22.6 & 2.1 & 0.9 & 73.7 & 75.0 \\
Strands-Decider-12B & Decision model & 41.4 & 0.0 & 2.0 & 40.4 & 61.1 \\
\bottomrule
\end{tabular}
}
\end{table}

\begin{table}[ht]
\centering
\caption{Results by task category (\%), over the twelve models in Table~\ref{tab:model_results} and three option orders.}
\label{tab:categories}
\resizebox{\linewidth}{!}{
\begin{tabular}{llccccc}
\toprule
\textbf{User} & \textbf{Category} & \textbf{Tasks} & \makecell{\textbf{Substeps}\\\textbf{per task}} & \makecell{\textbf{Mean step-wise}\\\textbf{accuracy}} & \makecell{\textbf{Best step-wise}\\\textbf{accuracy}} & \makecell{\textbf{Best task}\\\textbf{success}} \\
\midrule
Buyer & Gifts \& Discount Codes & 6 & 11.3 & 66.7 & 83.3 & 22.2 \\
Buyer & Order \& Delivery Changes & 7 & 9.9 & 52.4 & 71.0 & 9.5 \\
Buyer & Pre-purchase Product Questions & 11 & 10.1 & 53.8 & 70.6 & 3.0 \\
Buyer & Product Search \& Comparison & 12 & 10.3 & 59.7 & 75.8 & 8.3 \\
Buyer & Subscriptions \& Returns & 8 & 10.2 & 60.3 & 80.5 & 8.3 \\
Seller & Customer Questions \& Reviews & 8 & 10.5 & 49.5 & 74.6 & 4.2 \\
Seller & Fulfillment \& Shipping & 13 & 12.2 & 56.7 & 74.7 & 5.1 \\
Seller & Inventory \& Sales Analytics & 10 & 10.4 & 59.3 & 73.4 & 6.7 \\
Seller & Pricing Changes & 11 & 11.5 & 64.6 & 79.3 & 9.1 \\
Seller & Product Information Updates & 8 & 10.5 & 58.0 & 80.6 & 16.7 \\
Seller & Promotions & 6 & 11.8 & 69.4 & 86.9 & 11.1 \\
\bottomrule
\end{tabular}
}
\end{table}

\begin{figure}[ht]
\begin{promptwrapper}{Prompt for the backbone LLMs}
\small
\textbf{System:} You are the decision module of an AI agent. Read the state and choose the best option for the current step.

\medskip
\textbf{User:}\\
TASK: \{user query\}\\[2pt]
HISTORY:\\
\{substep\} \quad tool: \{reference tool\} \quad result: \{output, at most 400 characters\}\\
\ldots\\[2pt]
CURRENT STEP: \{current substep\}\\
Which tool should the agent call for the current step?\\
OPTIONS:\\
A. \{tool name\} -- \{tool description\}\\
\ldots\\
z. \{tool name\} -- \{tool description\}\\[2pt]
Answer with the option label only.
\end{promptwrapper}
\caption{Prompt for the backbone LLMs. The 52 candidates are labeled A--Z, then a--z.}
\label{fig:base_prompt}
\end{figure}

\clearpage

\section{Limitations and Ethical Considerations}
\label{app:limitations}

\textbf{Limitations.} \slate is synthetic: its queries, procedures, and tool schemas
are LLM-generated and its execution outcomes come from a simulator rather than live
APIs. This enables reproducible, large-scale evaluation but does not capture the
full behavioral complexity of real enterprise systems. Because generation used
Claude models, our study evaluates non-Claude, open models to avoid rewarding
generator-family bias; a sweep over larger and closed models is left to future work.
Our results cover six open models up to 12B parameters evaluated on \slate
itself---we do not report cross-benchmark or cross-domain comparisons here. They
isolate API selection by supplying reference arguments to correctly selected tools;
argument generation is supported by the harness but not evaluated here. Some substep
descriptions name their reference tool (16.1\% of tool steps), although accuracy on these
substeps is not higher than on the others (Table~\ref{tab:errors}).
The benchmark is instantiated in e-commerce; while
the construction pipeline is domain-agnostic, validating difficulty in other
domains (e.g., IT operations, finance) remains future work. Finally, runtime call
variations are matched to recorded entries by fixed normalization rules, which may
reject some valid paraphrases of free-form argument values.

\textbf{Ethical considerations.} \slate contains only synthetic data and no
personally identifying information. It is released for research on evaluating and
improving tool-use agents; as with any automation benchmark, deployment of agents
trained or selected against it should include human oversight for consequential
actions.

\end{document}